\documentclass[lettersize,journal]{IEEEtran}
\usepackage{amsmath,amsfonts}
\usepackage{algorithmic}
\usepackage{algorithm}
\usepackage{array}
\usepackage[caption=false,font=normalsize,labelfont=sf,textfont=sf]{subfig}
\usepackage{textcomp}
\usepackage{stfloats}
\usepackage{url}
\usepackage{verbatim}
\usepackage{graphicx}
\usepackage{cite}
\usepackage{color}
\newcommand*{\dif}{\mathop{}\!\mathrm{d}}
\usepackage{makecell}

\begin{document}

\title{Text2NeRF: Text-Driven 3D Scene Generation \\ with Neural Radiance Fields}

\author{
Jingbo~Zhang, Xiaoyu Li, Ziyu~Wan, Can Wang, and~Jing~Liao$^*$ 
\thanks{$^*$: corresponding author.} 
\thanks{J. Zhang, Z. Wan, C. Wang and J. Liao are with Department of Computer Science, City University of Hong Kong. E-mail: jbzhang6-c@my.cityu.edu.hk, ziyuwan2-c@my.cityu.edu.hk, cwang355-c@my.cityu.edu.hk, jingliao@cityu.edu.hk. X. Li is with Tencent AI Lab. E-mail: xliea@connect.ust.hk} 
}


\maketitle

\begin{abstract}
Text-driven 3D scene generation is widely applicable to video gaming, film industry, and metaverse applications that have a large demand for 3D scenes. However, existing text-to-3D generation methods are limited to producing 3D objects with simple geometries and dreamlike styles that lack realism. In this work, we present Text2NeRF, which is able to generate a wide range of 3D scenes with complicated geometric structures and high-fidelity textures purely from a text prompt. To this end, we adopt NeRF as the 3D representation and leverage a pre-trained text-to-image diffusion model to constrain the 3D reconstruction of the NeRF to reflect the scene description. Specifically, we employ the diffusion model to infer the text-related image as the content prior and use a monocular depth estimation method to offer the geometric prior. Both content and geometric priors are utilized to update the NeRF model. To guarantee textured and geometric consistency between different views, we introduce a progressive scene inpainting and updating strategy for novel view synthesis of the scene. Our method requires no additional training data but only a natural language description of the scene as the input. Extensive experiments demonstrate that our Text2NeRF outperforms existing methods in producing photo-realistic, multi-view consistent, and diverse 3D scenes from a variety of natural language prompts. Our code and model will be available upon acceptance.
\end{abstract}


\section{Introduction}

\IEEEPARstart{R}{ecent} breakthroughs in text-to-image generation have also sparked great interest in zero-shot text-to-3D generation \cite{jain2022zero, poole2022dreamfusion, wang2022score, seo2023let}, as using natural language prompts to specify desired 3D models is intuitive and, therefore, could increase the productivity of the 3D modeling workflow and reduce the entry barrier for novices. However, contrary to the text-to-image case, in which paired data is abundant, it is impractical to acquire large quantities of paired text and 3D data, making the text-to-3D generation task still challenging \cite{poole2022dreamfusion, ye20213d, lin2022magic3d}.

To circumvent this data limitation, some pioneer works, including CLIP-Mesh~\cite{khalid2022clip}, Dream Fields~\cite{jain2022zero}, DreamFusion~\cite{poole2022dreamfusion}, and Magic3D \cite{lin2022magic3d}, use deep priors of pre-trained text-to-image models, such as CLIP~\cite{radford2021learning} or image diffusion model~\cite{rombach2022high,saharia2022photorealistic}, to optimize a 3D representation, which thus empowers text-to-3D generation without the need for labeled 3D data. Despite the great success of these works, their generation results are still limited to 3D scenes with simple geometries and dreamlike styles. These limitations potentially stem from the fact that the deep priors derived from pre-trained image models, which are utilized to optimize the 3D representation, can only impose constraints on high-level semantics while neglecting low-level details. By contrast, recently concurrent arXived works, SceneScape \cite{fridman2023scenescape} and Text2Room \cite{hollein2023text2room}, directly employ the color image generated by text-image diffusion model to guide the reconstruction of 3D scenes. Although they support the generation of realistic 3D scenes, these methods mainly focus on indoor scenes and are hard to be extended into large-scale outdoor scenes due to the limitation of the explicit 3D mesh representation such that the stretched geometry caused by naive triangulation and noisy depth estimation. In contrast, our method utilizes NeRF as the 3D representation which has more advantage of modeling diverse scenes with complex geometry.

In this paper, we present Text2NeRF, a text-driven 3D scene generation framework by combining the strengths of Neural Radiance Fields (NeRF) \cite{mildenhall2021nerf} and text-image diffusion models. We adopt NeRF as our chosen 3D representation due to its superiority in capturing fine-grained and photorealistic details across a wide range of scenes~\cite{martin2021nerf,barron2022mip,deng2022fov}. This choice helps significantly suppress the artifacts caused by a triangular mesh representation, particularly in regions where depth discontinuity occurs. In addition, we use a pre-trained text-to-image diffusion model as the image-level prior to constrain the NeRF optimization from scratch without the demand of additional 3D supervision or multi-view training data.
Unlike the previous methods, e.g. DreamFusion \cite{poole2022dreamfusion}, that supervise the 3D generation with the semantic priors, we leverage finer-grained image priors inferred from the diffusion model, which consequently allows our Text2NeRF to generate more delicate geometric structure and realistic texture in the 3D scenes. Specifically, we employ the diffusion model to generate text-related images as the content prior and employ a monocular depth estimation method to offer the geometric prior of the generated scene. Both content and depth priors are leveraged to optimize the parameters of the NeRF representation. 

\begin{figure*}[!t]
  \centering
  \includegraphics[width=1\linewidth]{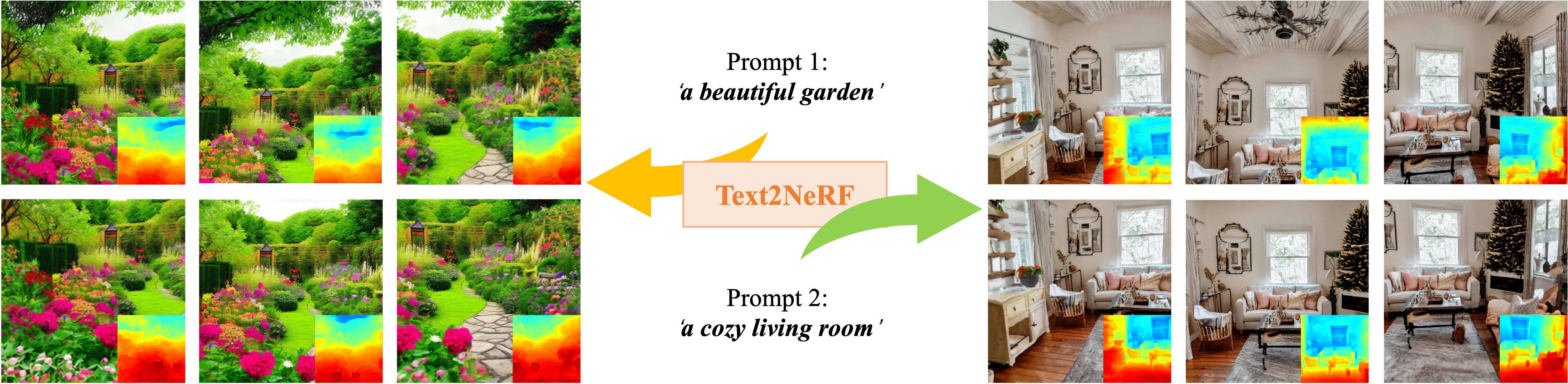}
  \caption{We propose Text2NeRF, a text-driven 3D scene generation framework by combining the neural radiance field representation and a pre-trained text-to-image diffusion model. Our Text2NeRF is capable of generating diverse and view-consistent indoor/outdoor 3D scenes solely from natural language descriptions. Please refer to our supplementary video demo for more examples.}
  \label{fig:teaser}
\end{figure*}

Moreover, to guarantee consistency between different views, we propose a progressive inpainting and updating strategy (PIU) for the novel view synthesis of 3D scenes. Through the PIU strategy, the generated scene can be expanded and updated in a view-by-view manner following a camera trajectory. 
In this way, the expanded area of the current view can be reflected in the next view by rendering the updated NeRF, which ensures that the same area will not be expanded repeatedly during the scene expansion process, thereby ensuring the continuity and view-consistency of the generated scene.
Briefly, the 3D representation of NeRF together with our PIU strategy ensures the view-consistent images generated by the diffusion model, resulting in a view-consistent 3D scene generation. 
In practice, we find that single-view training in NeRF leads to overfitting to that specific view, causing geometric ambiguity during view-by-view updating due to the lack of multi-view constraints. To overcome this issue, we construct a support set for the generated view, providing multi-view constraints for the NeRF modeling. 
Drawing inspiration from \cite{roessle2022dense}, in addition to image RGB loss, we also adopt an $L_2$ depth loss to achieve depth-aware NeRF optimization and improve the convergence rate and stability of NeRF models. Considering that the depth maps at different views are estimated independently and could be inconsistent in the overlapped regions, we further introduce a two-stage depth alignment strategy to align the depth values of corresponding points across different views, ensuring depth consistency. Thanks to these well-designed components, our Text2NeRF is capable of generating diverse, high-fidelity, and view-consistent 3D scenes solely from natural language descriptions, as shown in Fig.~\ref{fig:teaser}. The generality of our method allows for the generation of a wide range of 3D scenes, including indoor, outdoor, and even artistic scenes (Fig.~\ref{fig:more_results} and \ref{fig:more_results_cg}). Moreover, our approach is not limited by the view range and can generate 360-degree scenes (Fig.~\ref{fig:garden360}). Extensive experiments demonstrate that Text2NeRF outperforms previous methods both qualitatively and quantitatively.

Our contributions are summarized as follows:
\begin{itemize}
	\item We propose a text-driven realistic 3D scene generation framework combining diffusion models with NeRF, which supports zero-shot generation of various indoor/outdoor scenes from natural language prompts.
	\item We introduce the PIU strategy to progressively generate view-consistent novel contents for 3D scenes, and build the support set to provide multi-view constraints for the NeRF model during view-by-view updating.
	\item We employ the depth loss to achieve depth-aware NeRF optimization, and introduce a two-stage depth alignment strategy to eliminate estimated depth misalignment in different views.
\end{itemize}

\section{Related Work}
\label{sec:related}

\subsection{Text-Driven 3D Generation}
The long-standing problem of 3D generation entails constructing diverse view-consistent 3D geometry and high-fidelity textures. 
Early works, like 3D-GAN \cite{wu2016learning}, Pointflow \cite{yang2019pointflow}, and ShapeRF \cite{cai2020learning} focus more on the category-specific texture-less geometric shape generation based on the representations of voxels or point clouds. Subsequently, PlatonicGAN \cite{henzler2019escaping}, HoloGAN \cite{nguyen2019hologan}, and VolumeGAN \cite{xu20223d} are proposed to generate textured 3D scenes by learning the structural and textual representations from a category-specific dataset such as cars, faces, indoor scenes, et al. Although such methods achieve yield promising 3D scenes on specific categories, they cannot handle text-driven generative tasks. To achieve text-driven 3D generation, Text2shape \cite{chen2018text2shape} uses two encoder networks to learn cross-modal connections between texts and 3D models in the embedding space from a specific paired scene-text dataset. 

Thanks to the rapid development of text-to-image methods, recent works aim to employ the pre-trained text-to-image model to guide the 3D scene generation. For example, CLIP-Mesh \cite{khalid2022clip} adopts a semantically supervised optimization strategy to deduce shapes and textures for 3D meshes under the guidance of a pre-trained CLIP \cite{radford2021learning} model. Similar to CLIP-Mesh, PureCLIPNeRF \cite{lee2022understanding} and DreamFields \cite{jain2022zero} use the pre-trained CLIP model to guide the generation of 3D objects with implicit NeRF representations. Compared with the CLIP model, the state-of-the-art text image diffusion models \cite{ramesh2021zero, nichol2021glide, saharia2022photorealistic, rombach2022high} undoubtedly have more powerful generation capabilities due to their abundant training data and excellent structure. Therefore, DreamFusion \cite{poole2022dreamfusion} and SJC \cite{wang2022score} propose a score distillation sampling (SDS) loss to extract deep semantic priors from pre-trained text-to-image diffusion models \cite{saharia2022photorealistic, rombach2022high} and supervise the generative network of 3D models. Subsequently, some follow-up works, such as Magic3D \cite{lin2022magic3d}, Latent-NeRF \cite{metzer2022latent}, and 3DFuse \cite{seo2023let}, are proposed to improve the quality of generated 3D models under the constraint of SDS loss. Although these methods enable producing diverse 3D models related to the input prompts, they fail to generate a photorealistic 3D scene with complex geometry and high-fidelity textures because only high-level semantic priors of the pre-trained model are used to constrain the 3D generation. In contrast, our method infers low-level content and depth priors from the pre-trained text-to-image diffusion model, with which geometry and texture details in a photorealistic 3D scene are well constrained.

More recently, SceneScape \cite{fridman2023scenescape} and Text2Room \cite{hollein2023text2room}, which are \textit{independent and concurrent} to our work, propose text-to-3D schemes similar to our method. Differently, they employ explicit polygon meshes as the 3D representation during their generative procedure, which limits the representation of outdoor scenes and leads to stretched geometry and blurry artifacts in the fusion regions of mesh faces. In contrast, our implicit NeRF representation and reconstruction strategy could model fine-grained geometry and textures without specific scene requirements thus enabling our method to produce both indoor and outdoor scenes.

\subsection{Novel View Synthesis from a Single Image}
Some novel view synthesis methods constrained by 3D presentation are able to generate a 3D-consistent experience from a single image. For example, several existing 3D photography methods, like SVS \cite{tucker2020single}, 3DP \cite{shih20203d}, and 3D-Ken-Burns \cite{niklaus20193d}, use multi-plane images (MPI) or layered depth images (LDI) as 3D representations, and then employ pre-trained inpainting models to complete occluded regions to synthesize plausible novel views. However, such methods can only produce views in a small range due to the limitation of their specific 3D representation. By contrast, some other methods achieve the 3D reconstruction and novel view synthesis by mapping single-view image information to conventional 3D models. For instance, SynSin \cite{wiles2020synsin} transforms the image features into a point cloud based on the predicted depth information and decodes the rendered feature map to synthesize a novel view of the 3D scene. PixelSynth \cite{rockwell2021pixelsynth} constructs a point cloud by directly mapping the pixel color to the 3D points and introduces outpainting and refinement modules to fill the missing information in novel views. Worldsheet \cite{hu2021worldsheet} synthesizes novel views of the 3D scene by warping a planar mesh sheet according to the input image and predicted depth. Intuitively, directly applying one of these methods to extrapolate an image generated by a text-to-image model to novel views is a naive strategy for a text-driven 3D generation. However, this naive strategy is limited in several aspects. First, their scene extrapolation is based on the input image only, not conditioned on the text prompt. Consequently, their generated scene is within a limited view range around the input image to ensure semantic consistency. In contrast, our method allows for generating new content in novel views driven by the given text prompt. Therefore, ours is not limited by the view range and can even generate 360-degree scenes that are coherent with the text description. Besides, the explicit 3D representations, such as coarse mesh or point cloud, adopted in these methods restrict them from rendering fine results, while ours leveraging the implicit NeRF representation is superior in representing and rendering high-fidelity details.

\section{Method}
\label{sec:method}
\begin{figure*}[!t]
  \centering
  \includegraphics[width=1\linewidth]{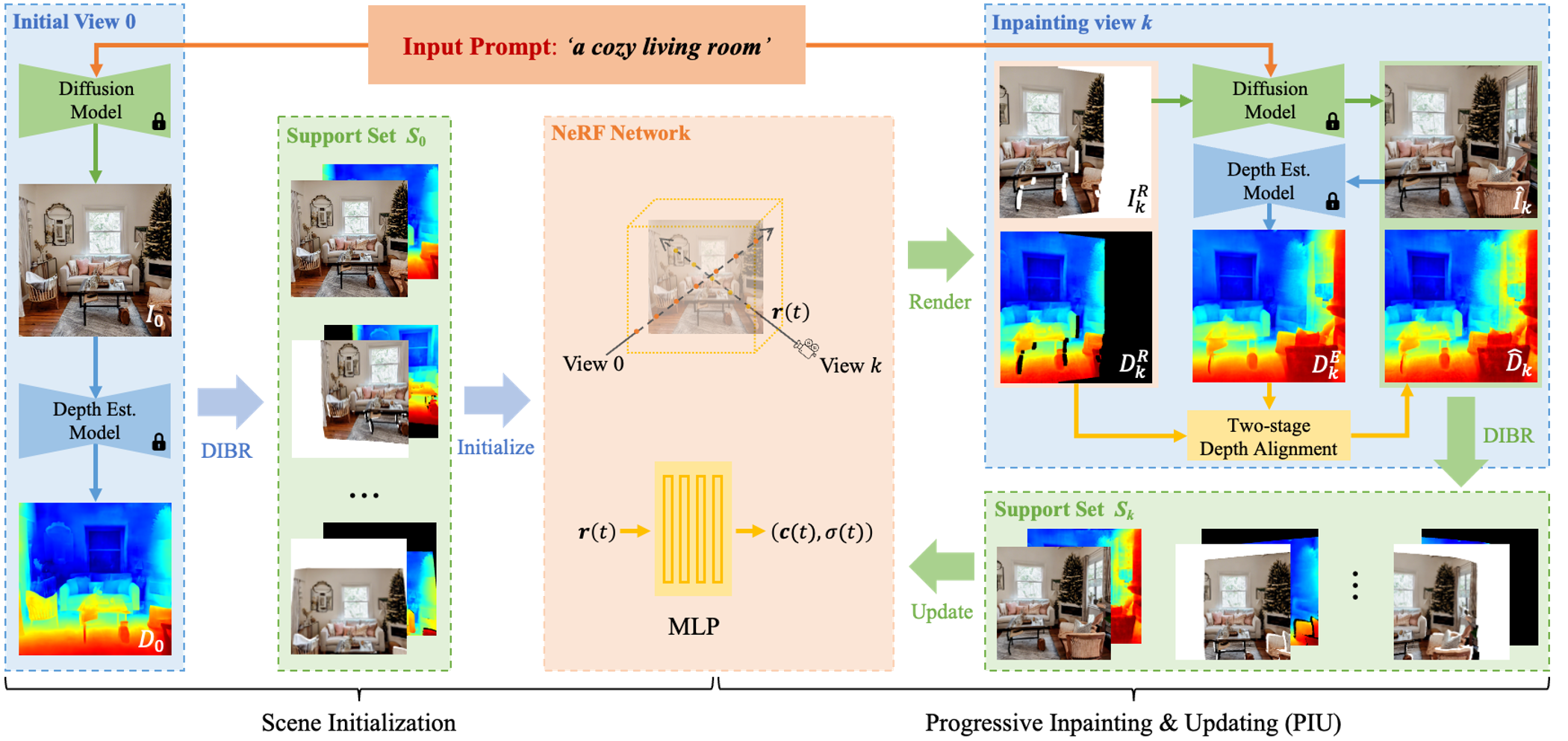}
  \caption{Overview of our Text2NeRF. Given an input text prompt, we infer an initial view $I_0$ and estimate its depth $D_0$ via a pre-trained diffusion model and a depth estimation model. Then we use the depth image-based rendering (DIBR) to warp the initial view and its depth map to various views to build the support set $\mathbf{S}_0$ for initializing the neural radiance field (NeRF). Afterward, we design a progressive inpainting and updating (PIU) strategy to complete missing regions consistently. During each update, we render the NeRF in a novel view $k$ to produce the image $I_k^R$ and depth $D_k^R$ with missing regions. Then, the diffusion model and depth estimation model are adopted to deduce completed image $\hat{I}_k$ and its depth $D_k^E$. Furthermore, a two-stage depth alignment is implemented on $D_k^R$ and $D_k^E$ to obtain aligned depth $\hat{D}_k$. Finally, the support set $\mathbf{S}_k$ of view $k$ is added into training data to update NeRF.}  
  \label{fig:pipeline}
\end{figure*}

We propose a text-driven 3D scene generation framework to progressively generate 3D scenes according to given text prompts as shown in Fig.~\ref{fig:pipeline}. We first generate an initial view by a text-to-image diffusion model. Based on the initial image, we build the support views and corresponding depth maps as the support set to offer multi-view constraints for the NeRF reconstruction using the depth image-based rendering (DIBR) method. After training this initialized NeRF model, we further introduce a progressive inpainting and updating (PIU) strategy to expand the generated scene view-by-view. Specifically, we render a novel view and complete its missing regions via the diffusion model with the text prompt. Then we take the inpainted view and constructed its support set as the additional supervision to update the NeRF model. By progressively adding new content consistent with the existing scene, our framework succeeds in generating realistic 3D scenes with fine-grained details.

\subsection{Scene Initialization}
\label{sec:init}
\noindent\textbf{Content Generation.} 
To obtain the initial scene content with respect to the input prompt $p$, we first employ a pre-trained diffusion model $f_{d}$ conditioned on $p$ to generate a 2D scene image $I_0=f_{d}(\epsilon \mid p)$, where $\epsilon$ is a random Gaussian noise. 
Due to the lack of geometric information in this single view $I_0$, a monocular depth estimation model $f_{e}$ is adopted to offer the geometric inference $D_0=f_{e}\left(I_0\right)$. The initial view $I_0$ and depth map $D_0$ will be used to produce a support set for the 3D scene initialization.

\noindent\textbf{3D Scene Representation.} 
Unlike explicit representations like polygon meshes or point clouds \cite{nehme2020visual, rockwell2021pixelsynth}, which are hard to represent complex geometry, NeRF shows its power in representing arbitrarily complex scenes. Therefore, we employ a NeRF network $f_\theta$ to represent the 3D scene. In NeRF, volume rendering \cite{mildenhall2021nerf} is used to accumulate the color in the radiance fields:
\begin{equation}
  \mathbf{C}(\mathbf{r}) = \int_{t_n}^{t_f}T(t) \sigma(\mathbf{r}(t)) \mathbf{c}(\mathbf{r}(t)) \dif t,
  \label{eq:volume_render}
\end{equation}
where $\mathbf{r}(t) = \mathbf{o} + t\mathbf{d}$ indicates the 3D coordinates of sampled points on the camera ray emitted from the camera center $\mathbf{o}$ with the direction $\mathbf{d}$.
$t_n$ and $t_f$ indicate the near and far sampling bounds. $(\mathbf{c}, \sigma) = f_\theta\left( \mathbf{r}(t) \right)$ are the predicted color and density of the sampled point along the ray. $T(t) =\exp(-\int_{t_n}^{t} \sigma(\mathbf{r}(s))\dif s)$ is the accumulated transmittance.
Different from vanilla NeRF that takes both 3D coordinate $\mathbf{r}(t)$ and view direction $\mathbf{d}$ in Eq.~\ref{eq:volume_render} to predict the radiance $\mathbf{c}$, we omit $\mathbf{d}$ to avoid the effect of view-dependent specularity.
Additionally, inspired by \cite{roessle2022dense}, we introduce the depth constraint into NeRF optimization to achieve depth-aware NeRF optimization and speed up model convergence. To this end, the predicted depth value $z(\mathbf{r})$ is required to be calculated:
\begin{equation}
  z(\mathbf{r}) = \int_{t_n}^{t_f}T(t) \sigma(\mathbf{r}(t)) t \dif t.
  \label{eq:depth_nerf}
\end{equation}
To be convenient, we denote the volume rendering on view $i$ as $(I_i^R, D_i^R)=VR(f_\theta \mid i)$, where $I_i^R$ and $D_i^R$ are the rendered image and depth map, respectively.

\noindent\textbf{Support Set.} Since the lack of multi-view supervision, directly adopting single-view $I_0$ and its depth $D_0$ to train the radiance fields easily leads to overfitting and geometric ambiguity. To overcome this issue, we adopt the depth image-based rendering (DIBR) method \cite{fehn2004depth} to construct a support set $\mathbf{S}_0$ for the initialization. Specifically, for each pixel $q$ in $I_0$ and its depth value $z$ in $D_0$, we compute its corresponding pixel $q_{0\rightarrow i}$ and depth $z_{0\rightarrow i}$ on a surrounding view $i$:
\begin{equation}
  \left[q_{0\rightarrow i}, z_{0\rightarrow i}\right]^{T} = \mathbf{K}\mathbf{P}_{i}\mathbf{P}_{0}^{-1}\mathbf{K}^{-1}\left[q, z\right]^{T}
  \label{eq:dibr}
\end{equation}
where $\mathbf{K}$ and $\mathbf{P}_{i}$ indicate the intrinsic matrix and the camera pose in view $i$. For convenience, we denote the DIBR process from view $0$ to view $i$ as $DIBR_{0\rightarrow i}$.

We manually set the intrinsic matrix $\mathbf{K}$ and camera pose $\mathbf{P}_{0}$ and then use $\mathbf{P}_{0}$ to get surrounding camera poses $\mathbf{P}_{i}$.
Specifically, we first define a surrounding circle of radius $\zeta$ centered at the current camera position and having the same z-coordinate as the current camera position. Then, we uniformly sample $\xi$ points as the camera positions and employ the same camera direction as the current view to produce the warping views in the support set. Here, $\zeta$ is the shift distance and $\xi$ is the number of warping views. In practice, we define $\xi=8$ by shifting the camera position with $\zeta=0.2$ in directions of up, down, left, right, upper left, lower left, upper right, and lower right, respectively. With these support views, along with the initial view $I_0$, we train a NeRF as the initialized 3D scene.

\subsection{Text-Driven Inpainting}
After the scene initialization, the radiance field can be rendered in arbitrary novel views. However, the rendered results other than the initial view $0$ will inevitably have missing content since the information in the initial scene is derived from the single image $I_0$. To complement the missing regions, we employ a text-driven inpainting method based on the pre-trained diffusion model $f_d$. Specifically, we first render a novel view $I_k^R$ to be inpainted. Then, we calculate the mask $M_k$ of missing parts in $I_k^R$ by warping all known views to the rendered view $k$ according to Eq.~\ref{eq:dibr}. The rendered image $I_k^R$ along with the mask $M_k$ and input prompt $p$ are fed into the diffusion model $f_d$ to predict an inpainting result of $I_k^R$:
\begin{equation}
  \hat{I}_k = f_d\left(I_k^R, M_k \mid p \right).
  \label{eq:inpaint}
\end{equation}

Considering that the inpainting process is stochastic, although the current diffusion model has a strong completion ability, it is difficult to guarantee that the quality of each result can meet the expected requirements. We thus perform the inpainting process many times for each view $I_k^R$ to be completed, and automatically select the one from all candidate inpainting results ${\hat{I}_k^j}$ that is most similar as the initial view in the CLIP semantic space:
\begin{equation}
  \hat{I}_k = \arg \max_{j} \cos{\left(E_{CLIP}\left(I_0\right), E_{CLIP}\left(\hat{I}_k^j\right)\right)},
  \label{eq:auto_select}
\end{equation}
where $E_{CLIP}(\cdot)$ is the image encoder of CLIP model \cite{wolf-etal-2020-transformers}. In practice, we generate 30 inpainting results as candidates for each view to be completed.

Besides, we also use the depth estimation model $f_e$ to estimate the depth map $D_k^E$ for $\hat{I}_k$. Note that, unlike the depth map $D_0$ of the initial view, $D_k^E$ cannot be directly taken as the supervision to update the radiance field since it is predicted independently and could conflict with known depth maps such as $D_k^R$ in the overlapping regions. To solve this issue, we implement depth alignment to align the estimated depth map to the known depth values in the radiance field.

\begin{figure}
  \centering
  \includegraphics[width=1\linewidth]{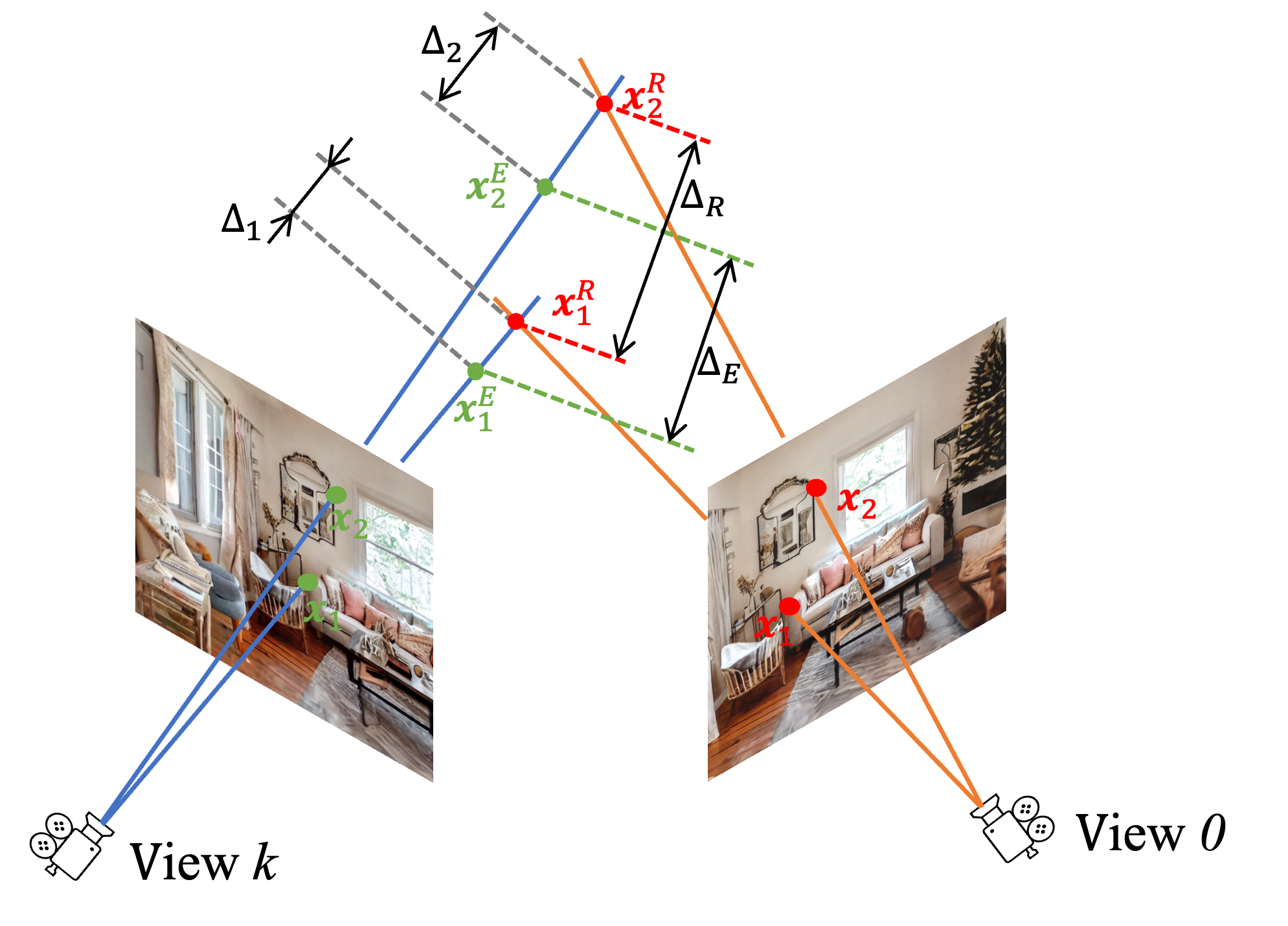}
  \caption{Example of scale and value differences. $x_1$ and $x_2$ are two aligned pixels in different views. The spacial points $x_1^E$ and $x_2^E$ are projected based on the estimated depth $D_k^E$ in view $k$. $x_1^R$ and $x_2^R$ are points projected according to the rendered depth $D_k^R$ which is constrained by known views such as view 0. Here, $\Delta_E \neq \Delta_R$ indicates the scale difference, and $\Delta_1 \neq 0$ or $\Delta_2 \neq 0$ indicate the value difference.}
  \label{fig:depth_align}
\end{figure}

\subsection{Depth Alignment}

Due to the lack of geometric constraint during the depth estimation, the predicted depth values could be misaligned in the overlapping regions \cite{luo2020consistent}, for example, the estimated depth $D_k^E$ of the inpainted view may be inconsistent with the depth $D_k^R$ rendered from NeRF since $D_k^R$ is constrained by previous known views.
The inconsistency is manifested in two aspects: scale difference and value difference. For instance, the \textit{distance difference} of two pixel-aligned spatial points and the \textit{depth value} of a specific point could be both different in depth maps estimated from different views, as shown in Fig.\ref{fig:depth_align}. The former is the scale difference and the latter is the value difference. In the case of scale difference, we cannot align both points by shift processing because even if we align the depth value of one of the points, the other point is still misaligned. To eliminate the scale and value differences between the overlapping regions of the rendered depth map $D_k^R$ and the estimated depth map $D_k^E$ of the novel view, we introduce a two-stage depth alignment strategy. Specifically, we first globally align these two depth maps by compensating for mean scale and value differences. Then we finetune a pre-trained depth alignment network to produce a locally aligned depth map. 

To determine the mean scale and value differences, we first randomly select $M$ pixel pairs from the overlapping regions and deduce their 3D positions under depth $D_k^R$ and $D_k^E$, denoted as $\left \{ (\mathbf{x}_j^R,\mathbf{x}_j^E) \right \}_{j=1}^M$. Next, we calculate the average scaling score $s$ and depth offset $\delta$ to approximate the mean scale and value differences:
\begin{equation}
  s = \frac{1}{M-1}\sum_{j=1}^{M-1}\frac{\Vert\mathbf{x}_j^R-\mathbf{x}_{j+1}^R\Vert_2}{\Vert\mathbf{x}_j^E-\mathbf{x}_{j+1}^E\Vert_2},
  \label{eq:scale}
\end{equation}
\begin{equation}
  \delta = \frac{1}{M}\sum_{j=1}^{M}\left(z\left(\mathbf{x}_j^R\right)-z\left(\mathbf{\hat{x}}_j^E\right)\right),
  \label{eq:offset}
\end{equation}
where $\mathbf{\hat{x}}_j^E = s \cdot \mathbf{x}_j^E$ indicates the scaled point and $z(\mathbf{x})$ represents the depth value of point $\mathbf{x}$. Then $D_k^E$ can be globally aligned with $D_k^R$ by $D_k^{global} = s \cdot D_k^E + \delta$.

Since depth maps used in our pipeline are predicted by a network, the differences between $D_k^R$ and $D_k^E$ are not linear, that is why the global depth aligning process cannot solve the misalignment problem. To further mitigate the local difference between $D_k^{global}$ and $D_k^R$, we train a pixel-to-pixel network $f_{\psi}$ for nonlinear depth alignment. During optimization of each view, we optimize the parameter $\psi$ of the pre-trained depth alignment network $f_{\psi}$ by minimizing their least square error in the overlapping regions:
\begin{equation}
  \mathop{\min}_{\psi} \left\| \left( f_{\psi}(D_k^{global}) - D_k^R \right) \odot M_k \right\|_2.
  \label{eq:local_align}
\end{equation}
Finally, we can derive the locally aligned depth using the optimized depth alignment network: $\hat{D}_k = f_{\psi}(D_k^{global})$. For convenience, we denote the two-stage depth alignment process as $align(D_k^E \mid D_k^R, M_k)$. In terms of the training of the depth alignment network, please refer to the implementation details in Sec.~\ref{sec_train}.

\begin{figure}[t!]
  \centering
  \includegraphics[width=1\linewidth]{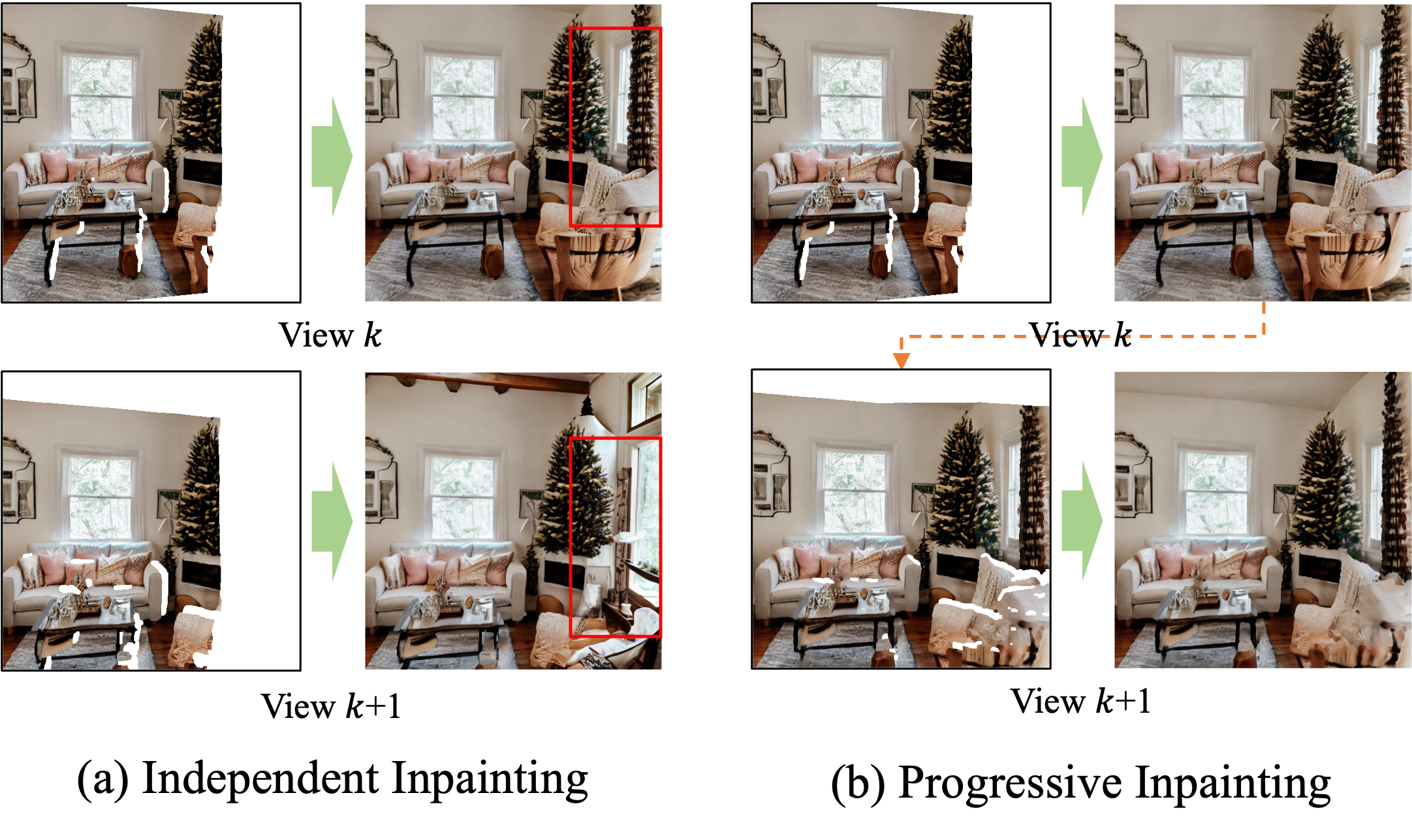}
  \caption{Examples of two inpainting strategies. The intuitive independent inpainting strategy simultaneously performs rendering and inpainting for each view. Due to there is no 3D constraint during 2D inpainting, the overlapping regions inpainted in different views will be view-inconsistent, as shown in the red box. In contrast, our progressive inpainting strategy achieves view-consistent inpainting results by introducing NeRF as a 3D constraint and reflecting previously inpainted content into the next view.}
  \label{fig:strategy_inpaint}
\end{figure}

\subsection{Progressive Inpainting and Updating}
After obtaining the inpainted image $\hat{I}_k$ and the aligned depth map $\hat{D}_k$ at iteration $k$, we could use Eq.~\ref{eq:dibr} to construct the corresponding support set $\mathbf{S}_k$ to update the radiance field. An intuitive solution is to render all the views from the initialized radiance field and inpaint them independently. However, there may be many overlapping regions to be inpainted among different views, so the 2D text-driven inpainting model cannot produce view-consistent content in all views without 3D constraints, as shown in Fig.~\ref{fig:strategy_inpaint}(a). To guarantee the view consistency and avoid the ambiguity of geometry and appearance during the scene inpainting process, we propose a progressive inpainting and updating strategy to update the radiance fields view by view, as shown in Fig.~\ref{fig:strategy_inpaint}(b) and Algorithm~\ref{alg:strategy}.
In this strategy, we update the radiance field $f_\theta$ after every inpainting process. It means that the previous inpainted content will be reflected in the subsequent renderings, and these parts will be regarded as known regions and will not be inpainted again in other views.

\begin{algorithm}[t!]
\caption{Progressive Inpainting \& Updating Strategy}
\label{alg:strategy}
\begin{algorithmic}[0]
  \renewcommand{\algorithmicrequire}{\textbf{Input:}} 
  \REQUIRE
    \STATE prompt $p$; 
    \STATE pre-trained diffusion model $f_d$;
    \STATE pre-trained depth estimation model $f_e$;
    \STATE initialized NeRF $f_\theta$;
    \STATE views to be updated $\mathbf{V} = \{1, 2, \cdots, N\}$;
    \STATE views already updated $\mathbf{\widetilde{V}} = \{0\}$.
  \renewcommand{\algorithmicrequire}{\textbf{Updating Process:}} 
  \REQUIRE
    \FOR{$k$ in $\mathbf{V}$}
        \STATE rendering $(I_k^R, D_k^R) = VR\left(f_\theta \mid k\right)$
        \STATE mask calculation $M_k\leftarrow\cap\{DIBR_{n\rightarrow k}\}$, where $n \in \widetilde{V}$
      \IF{$sum(M_k) > 0$}
        \STATE text-driven inpainting $\hat{I}_k = f_d\left(I_k^R, M_k \mid p \right)$
      \ELSE
        \STATE continue
      \ENDIF
        \STATE depth estimation $D_k^E=f_{e}(\hat{I}_k)$
        \STATE depth alignment $\hat{D}_k=align(D_k^E \mid D_k^R, M_k)$
        \STATE support set $\mathbf{S}_k\leftarrow\cup\{DIBR_{k\rightarrow support\_views}\}$
        \STATE update views updated $\mathbf{\widetilde{V}}= \mathbf{\widetilde{V}} \cup \{k\}$
        \STATE update NeRF model $f_\theta \leftarrow \mathbf{S}_k$
    \ENDFOR
  \renewcommand{\algorithmicensure}{\textbf{Return:}}
  \ENSURE updated NeRF $f_\theta$
\end{algorithmic}
\end{algorithm}

\subsection{Training and Implementation Details}
\label{sec_train}

\noindent\textbf{Training Objective.}
We use a RGB loss, a depth loss, and a transmittance loss to optimize the radiance field of the 3D scene. Like previous NeRF-based works\cite{mildenhall2021nerf, chen2022tensorf, song2023nerfplayer}, the RGB loss $L_{RGB}$ is defined as an $L_2$ loss between the rendered pixel color $\boldsymbol{C}^R$ and the color $\boldsymbol{C}$ generated by the diffusion model $f_d$. 
Different from previous works that employ regularized depth losses to handle uncertainty or scale-variant problem \cite{roessle2022dense, sargent2023vq3d}, we adopt a stricter depth loss $L_{Depth}$ to minimize the $L_2$ distance between the rendered depth $D^R$ and the aligned estimated depth $\hat{D}$, since the aligned depth maps used in our framework are scale-invariant and can be regarded as ground truth.
Besides, inspired by \cite{jain2022zero}, we design a depth-aware transmittance loss $L_T$ to encourage the NeRF network to produce empty density before the camera ray reaching the expected depth $\hat{z}$:
\begin{equation}
  L_{T} =  \Vert T(t)\cdot m(t) \Vert_2
  \label{eq:trans}
\end{equation}
where $m(t)$ is a mask indicator that satisfies $m(t)=1$ when $t<\hat{z}$, otherwise $m(t)=0$. $\hat{z}$ is the pixel-wise depth value in the aligned depth map $\hat{D}$. $T(t)$ is the accumulated transmittance which is same as the $T(t)$ in Eq.~\ref{eq:volume_render}. The total objective is then defined as:
\begin{equation}
  L_{total} =  L_{RGB} + \lambda_{d}L_{Depth} + \lambda_{t}L_{T},
  \label{eq:total}
\end{equation}
where $\lambda_{d}$ and $\lambda_{t}$ are constant hyperparameters balancing between terms.

\noindent\textbf{Implementation Details.}
We implement the Text2NeRF with the Pytorch framework \cite{paszke2019pytorch} and adopt TensoRF \cite{chen2022tensorf} as the radiance field. Note that, to make TensoRF satisfy the scene generation in a large view range, we let the camera position near the center of the NeRF bounding box and set outward-facing viewpoints. 
For scene generation, we use the stable diffusion model in version 2.0 \cite{rombach2022high} to generate the scene content related to the input prompt and use the boosting monocular depth estimation method \cite{miangoleh2021boosting} with pre-trained \textit{LeReS} model \cite{yin2021learning} to estimate the depth for each view.
In term of depth alignment, the super-parameter $M$ in Eq.~\ref{eq:scale} is set as $\min(M_0, 10000)$ in practice, where $M_0$ indicates the number of all matched points in the overlapping regions. Besides, the depth alignment network in our framework uses the same pixel-to-pixel U-net architecture as the depth merging network in \cite{miangoleh2021boosting}. To train this network, we first predict 10000 depth maps using the depth estimation models and add continuous non-linear random noise into these depth maps, i.e., $\widetilde{D} = \left(D+\tau_1\right)\cdot D^{{1}/{\tau_2}}$ where $D$ is the depth; $\tau_1$ and $\tau_2$ indicate the shift and scale factors, which are randomly sampled in the range $[0,1]$ and $[30,50]$, respectively. Then, we use the noisy depth maps as input and constrain the depth alignment network with the noise-free depth maps, so that the network acquires the ability to locally change the depth value. Finally, we finetune the network based on Eq.~\ref{eq:local_align} to produce the local aligned depth for each inpainting view.
During training, we use the same setting as \cite{chen2022tensorf} for the optimizer and learning rate and set the hyperparameters in our objective function as $\lambda_{d}=0.005$ and $\lambda_{t}=1000$.

\section{Experiments}

\begin{figure*}
  \centering
  \includegraphics[width=1\linewidth]{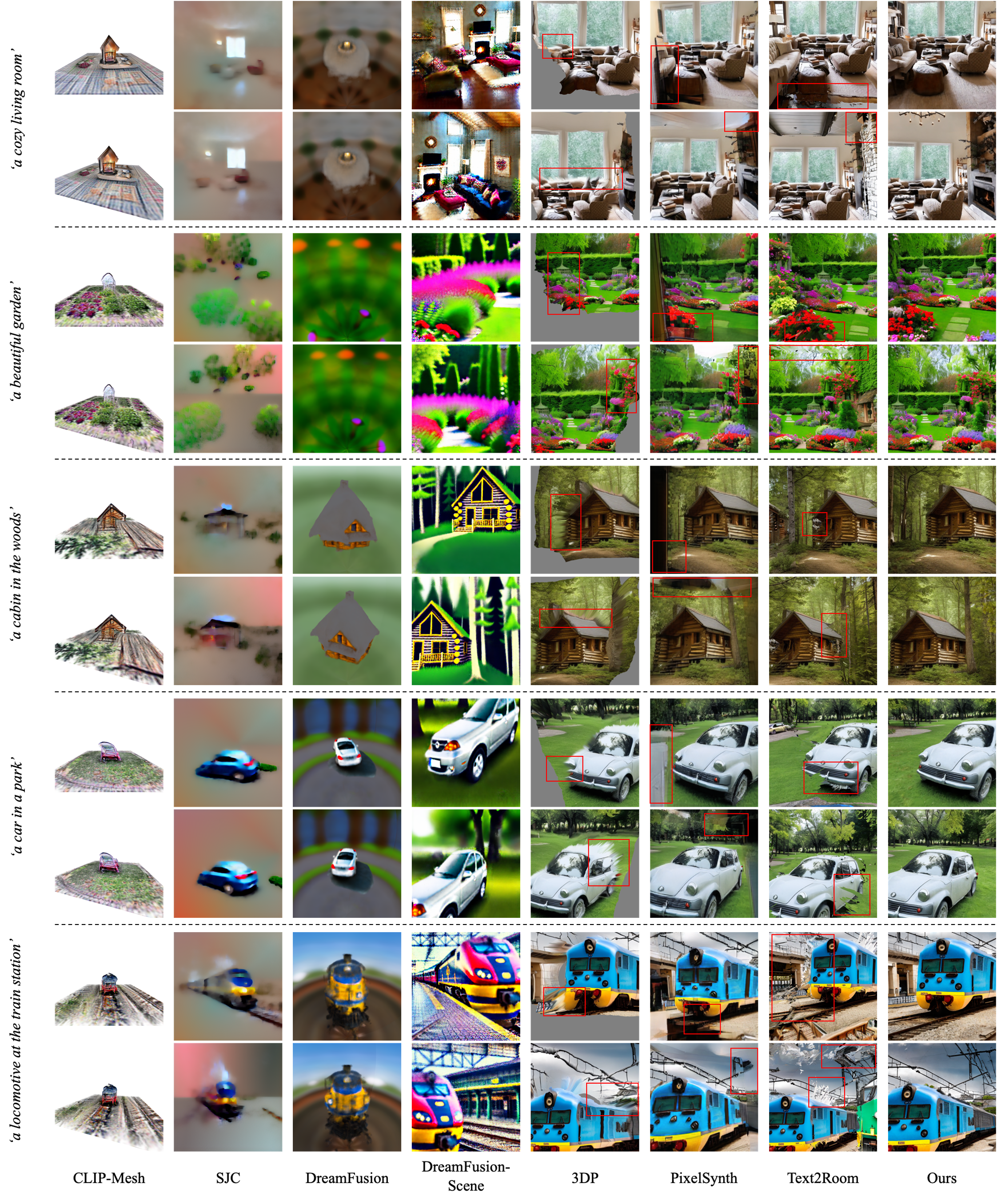}
  \caption{Qualitative comparison of results generated by baselines and ours on different text prompts. Here, we only show two rendering results from different views for each generated scene of each method due to space limitations. Please refer to the supplementary material for video results.}
  \label{fig:exp_all}
\end{figure*}

\begin{table*}[!t]
   \caption{Discrimination of baseline methods and ours in guidance type and 3D representation, optimization time, and quantitative comparison. Here, \textit{S} indicates high-level semantic prior and \textit{I} represents low-level image prior.
   Compared to baseline methods, our Text2NeRF yields a lower metric score on both BRISQUE and NIQE and a higher score on CLIP similarity, which means that our method achieves to generate more realistic and higher-quality text-related results.}
  \centering
  \begin{tabular}{@{}lcccccccc@{}}
    \hline
    Methods & CLIP-Mesh & SJC & DreamFusion & DreamFusion-Scene & 3DP & PixelSynth  & Text2Room & \textbf{Ours} \\
    \hline
    Guidance Type & S & S & S & S & I & I & I & I \\
    3D Representation & Mesh & NeRF & NeRF & NeRF & LDI\&Mesh & Point Cloud & Mesh & NeRF\\
    \hline
    Opti. time (hours) & 0.206 & 0.429 & 1.148 & 1.510 & 0.125 & 0.409 & 0.365 & 1.525 \\
    \hline
    BRISQUE $\downarrow$ & 46.266 & 39.543 & 67.012 & 37.799 & 30.592 & 25.924 & 28.395 & \textbf{24.498} \\
    NIQE $\downarrow$ & 6.652 & 11.971 & 12.022 & 6.402 & 6.260 & 6.604 & 5.415 & \textbf{4.618} \\
    CLIP Score $\uparrow$ & 27.480 & 24.152 & 22.576 & 28.032 & 27.376 & 27.267 & 28.056 & \textbf{28.695} \\
    \hline
    
  \end{tabular}
  \label{tab:exp_all}
\end{table*}

\label{sec:exp}
In this section, we first briefly introduce several state-of-the-art text-to-3D baselines and metrics (Sec.~\ref{sec:setup}), and then we apply our Text2NeRF to a variety of text prompts to evaluate its capability on photo-realistic indoor and outdoor 3D scenes generation and compare with the baseline methods~(Sec.~\ref{sec:results}). Furthermore, we conduct ablation studies to investigate the effectiveness of major components in our method (Sec.~\ref{sec:abla}). 

\subsection{Setup}
\label{sec:setup}
\noindent\textbf{Baseline Methods.}
To evaluate the performance of our method on text-driven 3D scene generation, we compare our method with seven baseline methods, as shown in Table.~\ref{tab:exp_all}, including four generation methods guided by the high-level semantic prior (i.e., CLIP-Mesh \cite{khalid2022clip}, SJC \cite{wang2022score}, DreamFusion \cite{poole2022dreamfusion}, and DreamFusion-Scene) and three methods guided by the low-level image prior (i.e., 3DP \cite{shih20203d}, PixelSynth \cite{rockwell2021pixelsynth}, and Text2Room \cite{hollein2023text2room}). Here, CLIP-Mesh, SJC, and DreamFusion are three existing state-of-the-art text-to-3D methods which employ NeRF as their 3D representation. DreamFusion-Scene is a modified version of DreamFusion designed for generating 3D scenes, as the vanilla version focuses on 3D objects and is not suitable for outward-facing scene generation. 3DP and PixelSynth are two novel view synthesis methods using explicit polygon meshes or point clouds as 3D representation, which represent a naive strategy for the text-driven 3D generation, i.e., applying existing novel view synthesis methods to the single image generated by a text-to-image diffusion model. Text2Room is one recently arXived concurrent work which employ polygon meshes to represent the generated 3D scenes.
Notably, due to DreamFusion being performed based on the unavailable Imagen \cite{saharia2022photorealistic} diffusion model, we replace it with a Pytorth implementation\footnote{https://github.com/ashawkey/stable-dreamfusion} powered by the stable diffusion \cite{rombach2022high} model.

\noindent\textbf{Metrics.}
Since there is no ground truth as a reference for generated 3D scenes related to the text prompts, previous reference-based metrics are not suitable for the generation tasks, like PSNR and LPIPS \cite{zhang2018unreasonable}. Instead, we use two metrics, blind/referenceless image spatial quality evaluator (BRISQUE) \cite{mittal2012no} and natural image quality evaluator (NIQE) \cite{mittal2012making}, on no-reference image quality assessment to evaluate rendering quality of generated 3D scenes. Besides, we adopt the CLIP text-image similarity score \cite{radford2021learning} to measure how well the rendered images align with the input prompt. 

\begin{figure*}[t]
  \centering
  \includegraphics[width=1\linewidth]{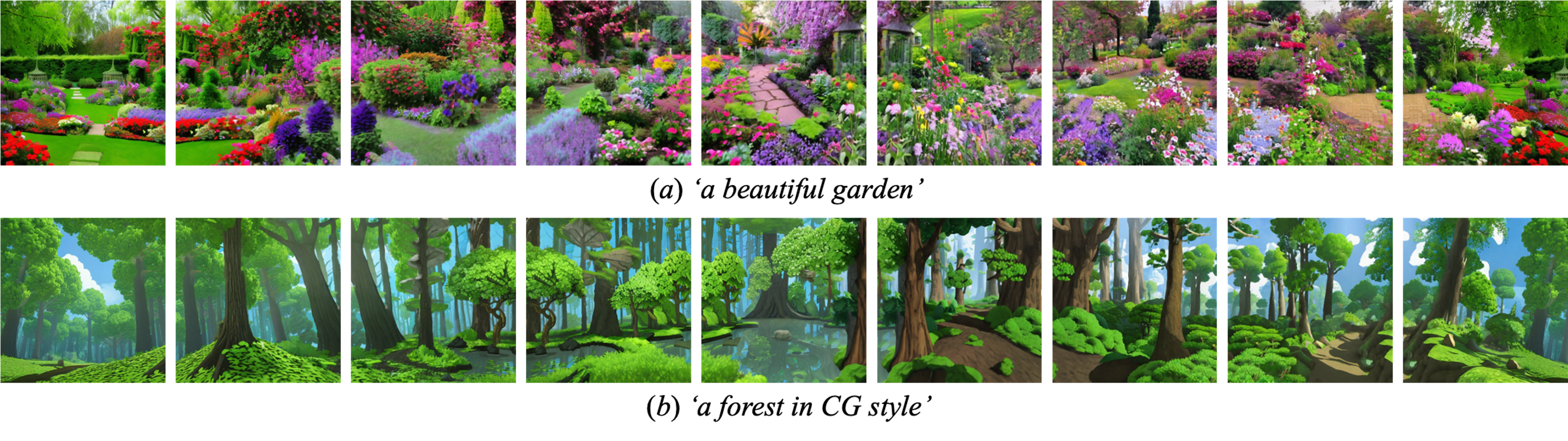}
  \caption{360-degree scenes generated by our Text2NeRF. Please refer to the supplementary material for video results.}
  \label{fig:garden360}
\end{figure*}

\begin{figure*}
  \centering
  \includegraphics[width=1\linewidth]{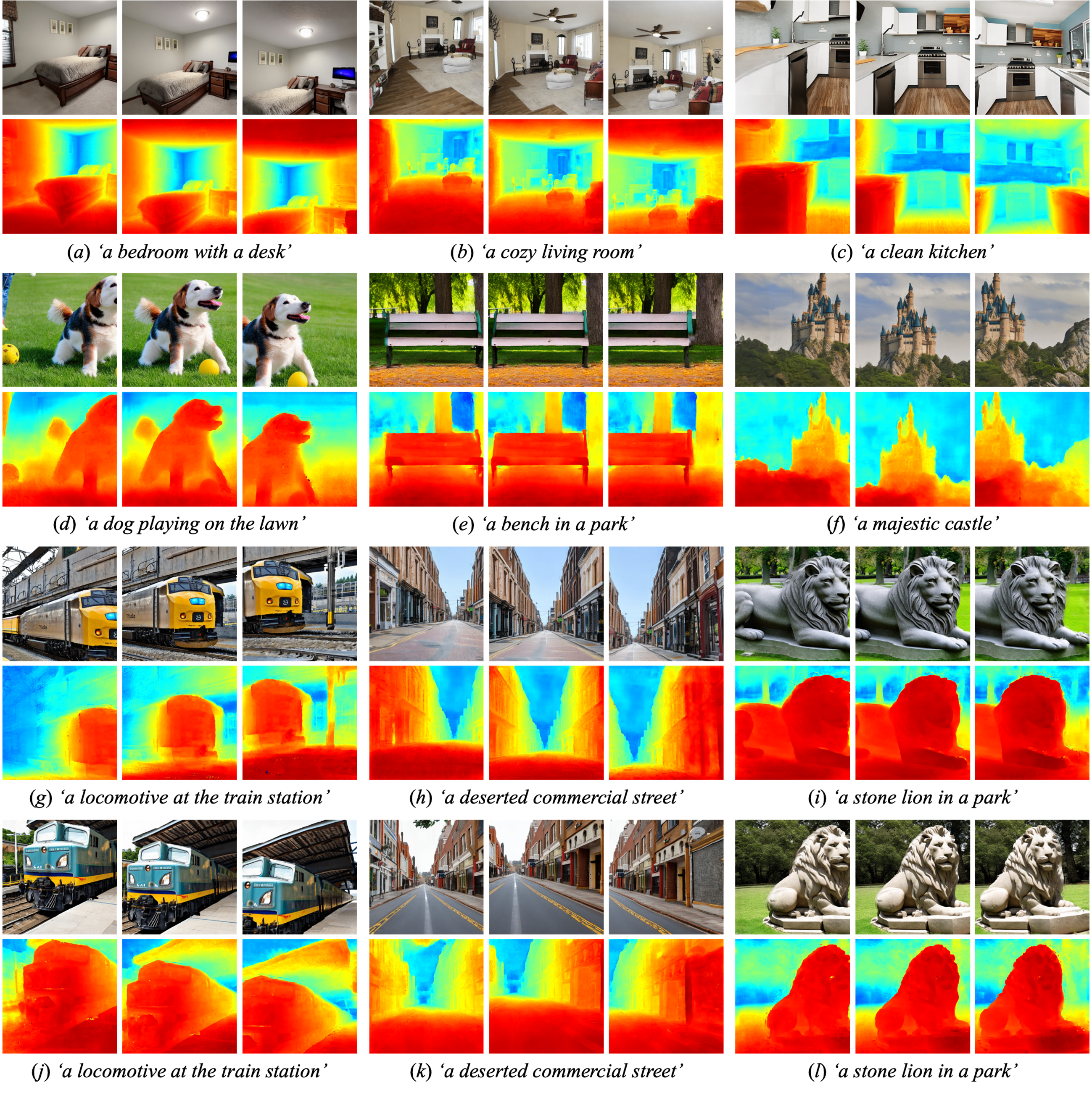}
  \caption{More results of our 3D scene generation. It is worth noting that our method can generate diverse results from the same text prompt (g)\&(j), (h)\&(k), and (i)\&(l). Please refer to the supplementary material for video results.}
  \label{fig:more_results}
\end{figure*}

\begin{figure*}
  \centering
  \includegraphics[width=1\linewidth]{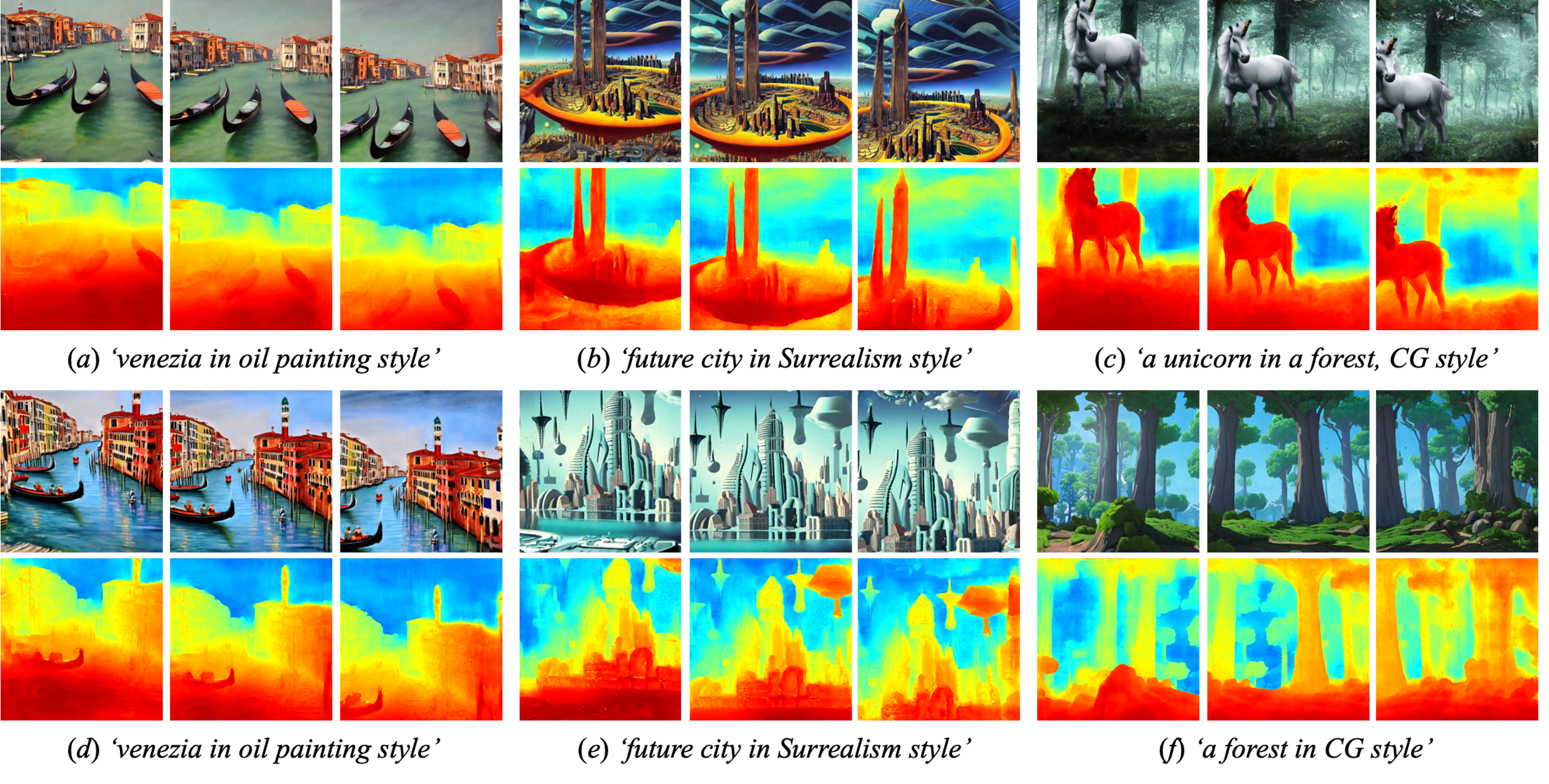}
  \caption{3D scenes in artistic styles generated by our Text2NeRF. Please refer to the supplementary material for video results.}
  \label{fig:more_results_cg}
  \vspace{-0.05in}
\end{figure*}

\subsection{Comparisons}
\label{sec:results}
We evaluate our Text2NeRF and compare it with baseline methods for text-driven 3D scene generation across various prompts, as shown in Fig.~\ref{fig:exp_all}. Additionally, we provide the average evaluation scores of BRISQUE, NIQE, and CLIP for the rendered images produced by different methods, as shown in Tab.~\ref{tab:exp_all}. Clearly, our method surpasses the baselines by generating higher-quality 3D scenes, as indicated by lower BRISQUE and NIQE values. Moreover, our method ensures the semantic relevance between the generated scene and the input text, resulting in a higher CLIP score. Overall, both qualitative and quantitative results unequivocally demonstrate the superiority of our approach over the baseline methods.

As shown in the first three columns of Fig.~\ref{fig:exp_all}, CLIP-Mesh, SJC, and DreamFusion struggle to generate complex 3D scenes related to the given prompts since their primary design focus on simple 3D object generation. Consequently, their BRISQUE and NIQE values tend to be higher compared to other methods, indicating relatively poorer quality in the rendered images of their generated scenes. In particular, CLIP-Mesh generates 3D scenes by optimizing initial sphere and planar multi-mesh representations, guided by a pre-trained CLIP model. Due to the absorption of environmental semantics into the planar mesh during optimization, CLIP-Mesh is limited to producing object-centric scenes. Similarly, SJC and DreamFusion adopt a \textit{looking-inside} camera setting and sample the camera position in outer spherical coordinates of the radiation field. In this way, the unbounded background environment is difficult to optimize in the central radiance field, resulting in the tendency of both SJC and DreamFusion to also generate object-centric scenes. Unlike SJC, DreamFusion incorporates an additional background spherical surface outside the central radiance field. This design choice allows DreamFusion to include the scene environment in the background representation, fulfilling high-level semantic priors, as observed in the examples of the \textit{garden} and \textit{car}.
Excluding completely failed cases, CLIP-Mesh, DreamFusion, and SJC exhibit the ability to generate object-centric scenes with a dreamlike style. However, they struggle to create 3D scenarios with complex spatial arrangements and geometry. In contrast, the modified DreamFusion-Scene successfully generates text-related 3D scenes with more complex geometry. Nevertheless, DreamFusion-Scene still falls short in deducing detailed structures and achieving photorealistic textures for the generated scenes. This limitation stems from the fact that the deep semantic priors provided by the text-image method are insufficient to fully constrain the low-level details.

Unlike existing text-to-3D methods guided by the deep semantic priors, the naive strategy that utilizes the novel view synthesis methods, 3DP and PixelSynth, to reconstruct the 3D scene from a single text-related image generated by the text-image model. The fifth and sixth columns of Fig.~\ref{fig:exp_all} demonstrate that such methods achieve to produce photo-realistic text-related 3D scenes with textual details, since they leverage the low-level content and depth priors to guide the 3D reconstruction process. As a result, their BRISQUE and NIQE values are substantially lower than those of previous semantic-guided generation methods, indicating superior scene quality and realism. However, their scene extrapolation is implemented within a limited view range and is independent of the input prompt, making it difficult for them to generate semantically consistent content in some novel views of the scene.
Specifically, 3DP employs LDI and polygon meshes to represent the reconstructed 3D scene, which is susceptible to depth discontinuities. This can lead to missing content or stretched geometry in regions where depth is discontinuous, as illustrated in the gray area and red box in the fifth column of Fig.~\ref{fig:exp_all}. By contrast, PixelSynth represents the 3D scene as point clouds, which mitigates the sensitivity to depth discontinuities to some extent. However, limited by its prompt-independent inpainting module, PixelSynth is prone to generating incoherent and blurry content, especially in the inpainted regions. Moreover, as shown in Fig.~\ref{fig:garden360}, our Text2NeRF supports text-driven scene generation in a large view range thanks to our progressive scene inpainting and updating strategy. On the other hand, other novel view synthesis methods produce blurred scene-filling results even at a small viewing angle since the text-related guidance is not considered in such methods.

In comparison to the novel view synthesis methods, both the concurrent work Text2Room and ours leverage the text-conditioned diffusion model as an inpainting module to complete missing regions in 3D scenes. To preserve the low-level textural details in the 2D images generated by the diffusion model, we both introduce a color objective as the low-level image guidance. This shared characteristic allows both approaches to generate 3D scenes that simultaneously exhibit high quality (as indicated by low BRISQUE and NIQE values) and high semantic relevance (as reflected in high CLIP scores). However, there are differences in how the generated scenes are represented. Unlike Text2Room that utilizes polygon meshes to represent the generated scenes, we adopt the NeRF (Neural Radiance Fields) framework, encoding the 3D scenes in an implicit network. This choice enables our method to effectively model fine-grained and photorealistic details in both bounded and unbounded scenes. As shown in the seventh column of Fig.~\ref{fig:exp_all}, Text2Room encounters challenges in generating certain outdoor scenes and often produces stretched geometry in regions with depth discontinuity. In contrast, our method successfully generates indoor and outdoor 3D scenes with complex structures and achieves a higher level of photorealistic details in depth discontinuous regions.

Furthermore, we show more examples of 3D scenes generated by our Text2NeRF in Fig.~\ref{fig:more_results}. It is worth noting that our method can not only generate diverse results from the same text prompt (Fig.~\ref{fig:more_results}(g)\&(j), (h)\&(k), and (i)\&(l)), but also support to generate 3D scenes in some artistic styles (Fig.~\ref{fig:more_results_cg}). Please refer to the supplementary material for video results.

\subsection{Ablation Studies}
\label{sec:abla}
\begin{figure}[t]
  \centering
  \includegraphics[width=1\linewidth]{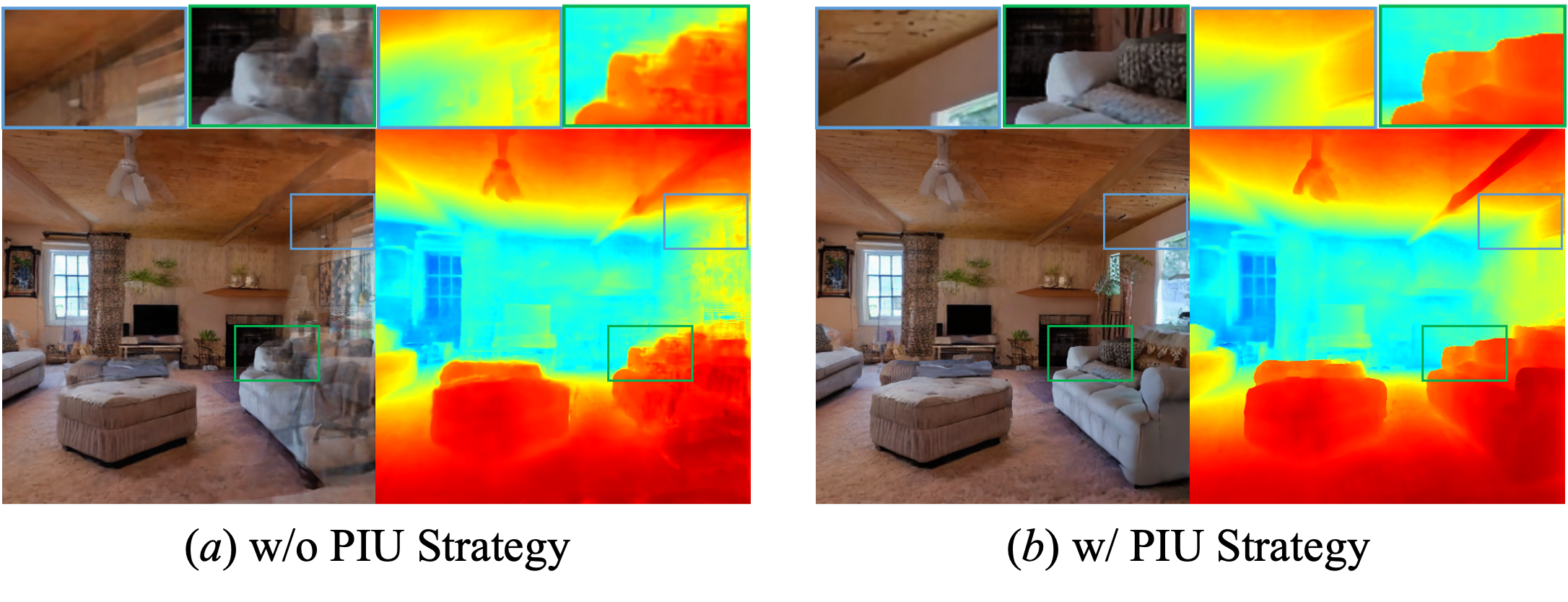}
  \caption{Effectiveness validation of the PIU strategy. In the absence of the PIU (Progressive Inpainting and Updating) strategy, the missing regions in different views are independently inpainted, leading to noticeable artifacts in the final generated scene. However, by incorporating the PIU strategy, the generated scene is inpainted and updated in a view-by-view manner, ensuring view consistency and producing 3D scenes with distinct textures.}
  \label{fig:abl_piu}
\end{figure}

\noindent\textbf{Ablation on PIU Strategy.}~
To investigate the effectiveness of the PIU strategy in our pipeline, we conduct a comparative experiment by replacing it with the independent inpainting strategy. As shown in Fig.~\ref{fig:abl_piu}, in the absence of the PIU strategy, view-inconsistent inpainted views provide equal constraints on the content of the radiation field, which in turn produces significant artifacts in overlapping regions. Accordingly, the BRISQUE and NIQE values in Tab.~\ref{tab:abla} are higher compared to those obtained by our full method. By contrast, our PIU strategy enables the generation process to proceed view by view, effectively avoiding the view-inconsistent problem that may occur in the completion area. 

\begin{figure}[t]
  \centering
  \includegraphics[width=1\linewidth]{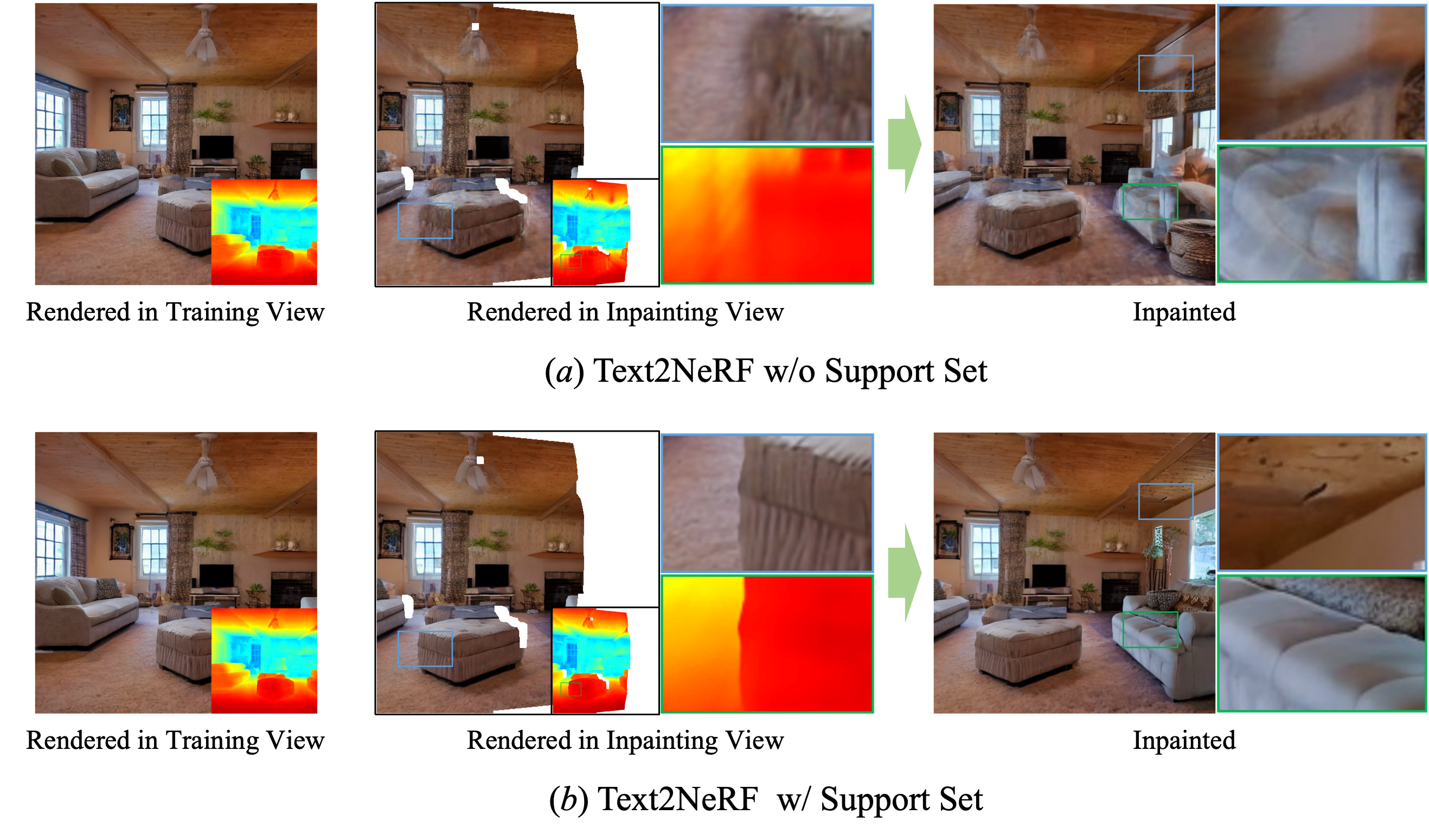}
  \caption{Effectiveness validation of support set. Without the support set, although NeRF achieves good rendered image in the training view due to overfitting, it cannot produce a clear result in a novel inpainting view. By contrast, the case with support set enable to obtain images with desired quality in both training and inpainting views. Correspondingly, compared to the blurry rendering image, the clear one contributes to a better inpainted result.}
  \label{fig:abl_sprt}
\end{figure}

\begin{figure}[t]
  \centering
  \includegraphics[width=1\linewidth]{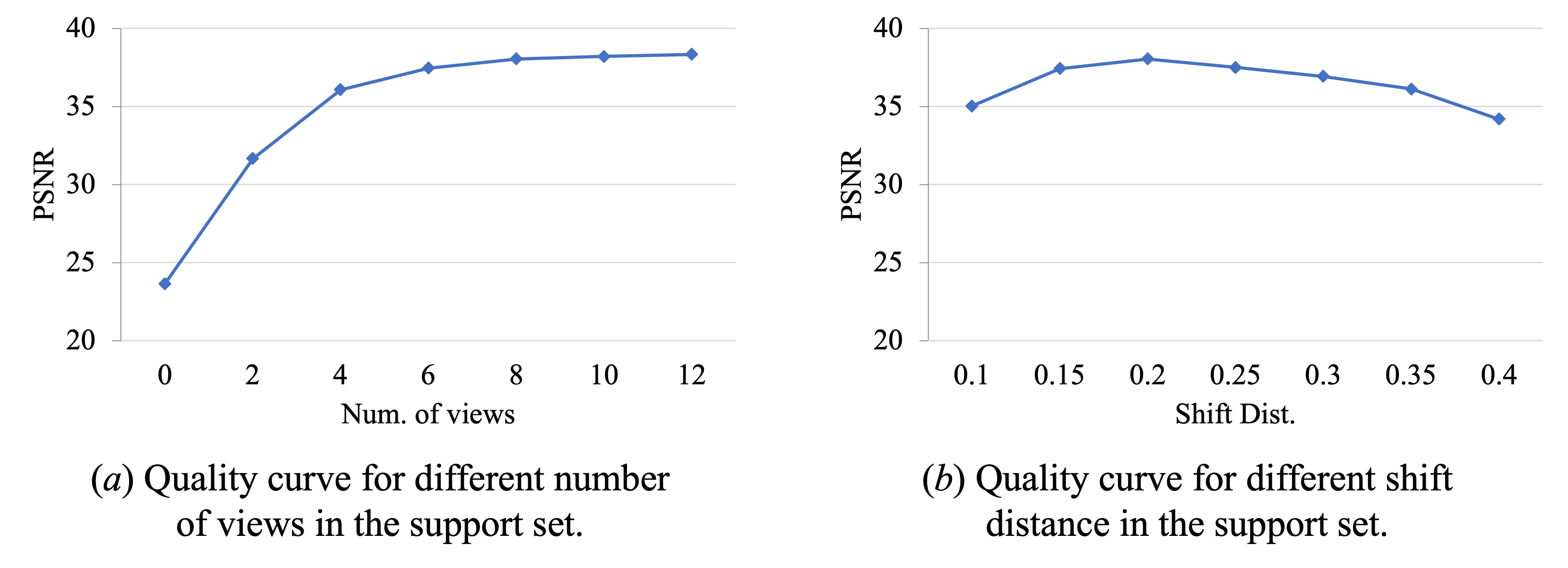}
  \caption{Quality curves for different number of warping views and shift distance in the support set. Note that number $0$ indicates the implementation without support set.}
  \label{fig:curve}
\end{figure}

\begin{figure}[t]
  \centering
  \includegraphics[width=1\linewidth]{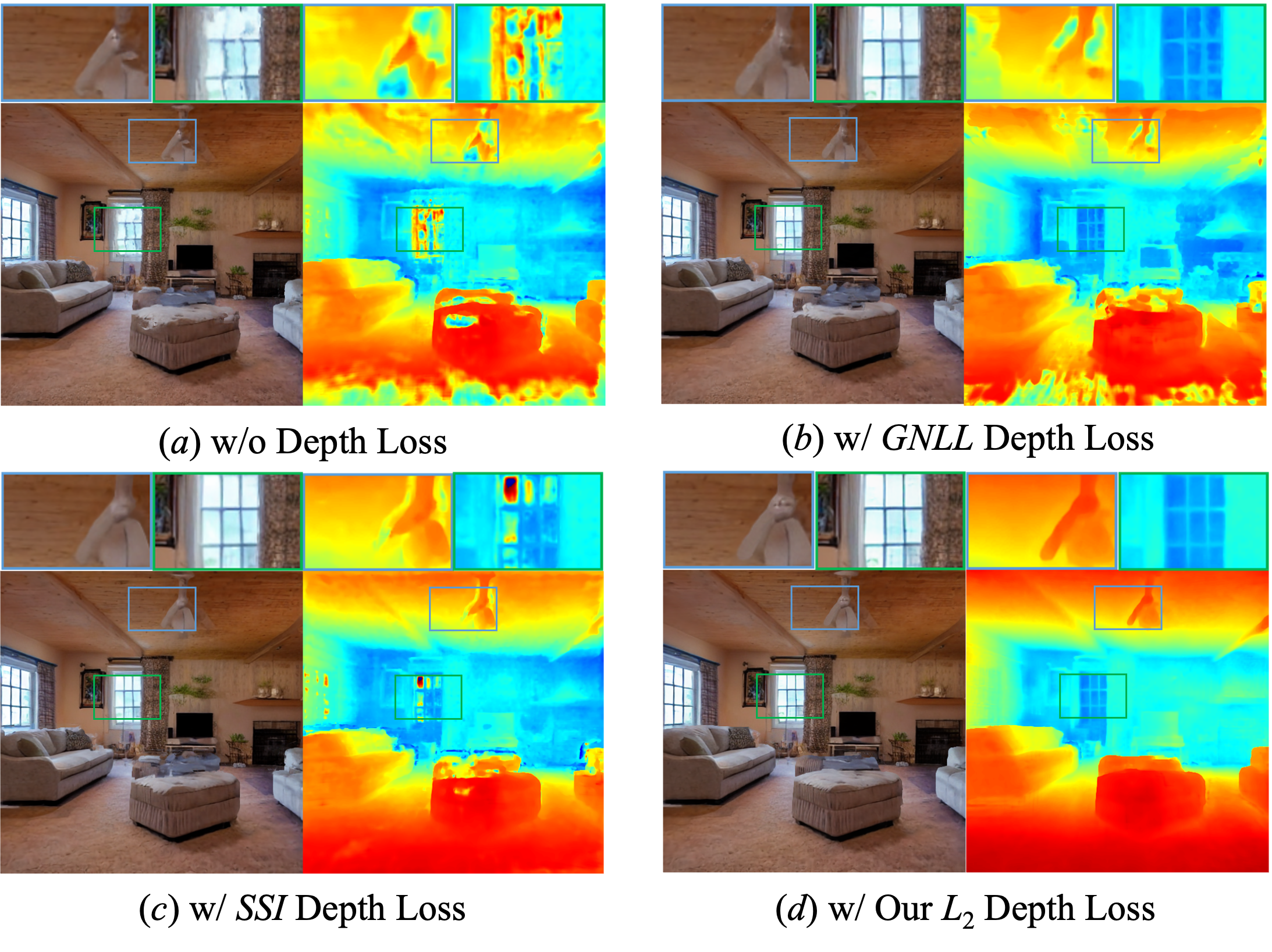}
  \caption{Effectiveness validation of our depth loss. Without the guidance of depth information, ambiguous depth values are produced in the near and far areas. In contrast, GNLL and SSI losses can constrain the depth values to a certain extent, but still cannot provide a strict constraint like our $L2$ depth loss.}
  \label{fig:abl_loss}
\end{figure}

\begin{figure}[t]
  \centering
  \includegraphics[width=1\linewidth]{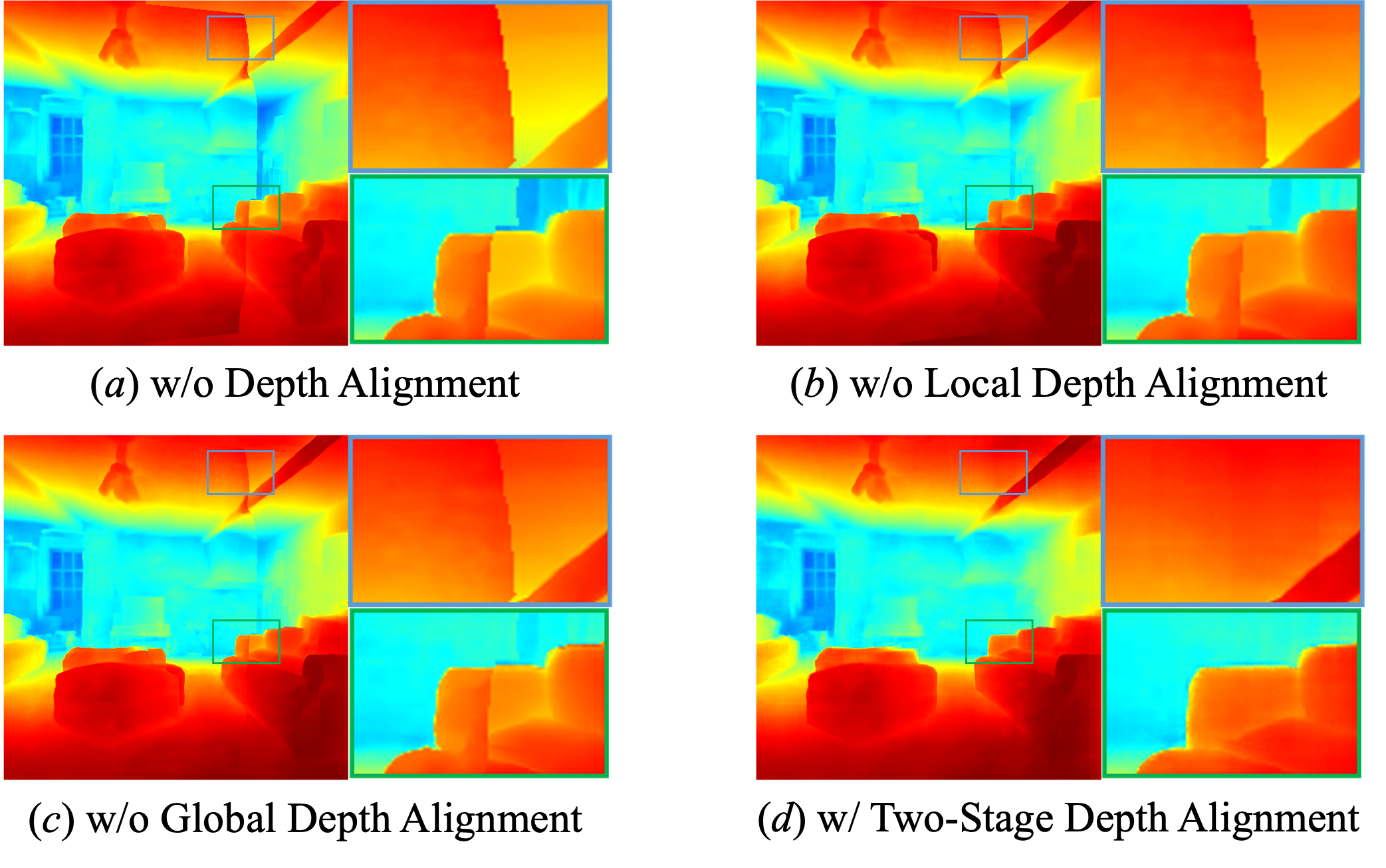}
  \caption{Effectiveness validation of our two-stage depth alignment. In the absence of depth alignment, prominent demarcation lines arise due to depth differences in the merged depth map. Global alignment helps bring the newly estimated depth values closer to the known depth map as a whole, but fails to eliminate the demarcation lines entirely. A comparison between (a) and (c) reveals that local alignment improves the alignment of unaligned depth maps, yet without global alignment, complete elimination of differences remains challenging. In contrast, our two-stage strategy effectively achieves smoother transitions and harmonious results at the demarcation lines.}
  \label{fig:abl_align}
\end{figure}

\begin{table*}[!t]
   \caption{Ablation studies on key components of our method.}
  \centering
  \begin{tabular}{@{}lccccccccc@{}}
    \hline
    Methods & \thead{Ours \\ (full)} & \thead{w/o \\ PIU} & \thead{w/o \\ Support Set} & \thead{w/o \\ Depth Loss} & \thead{w/ \textit{GNLL} \\ Depth Loss} & \thead{w/ \textit{SSI} \\ Depth Loss} & \thead{w/o \\ Depth Align.} & \thead{w/o Local \\ Depth Align.} & \thead{w/o Global \\ Depth Align.}\\
    \hline
    BRISQUE $\downarrow$ & \textbf{24.498} & 33.8434 & 28.3389 & 27.0617 & 26.1802 & 25.8995 & 28.1771 & 27.9174 & 27.1946 \\
    NIQE $\downarrow$ & \textbf{4.618} & 6.012 & 5.778 & 5.588 & 4.837 & 4.711 & 5.945 & 5.209 & 4.839 \\
    CLIP Score $\uparrow$ & \textbf{28.695} & 25.733 & 26.330 & 26.782 & 27.168 & 27.126 & 26.173 & 26.527 & 26.811 \\
    \hline
  \end{tabular}
  \label{tab:abla}
\end{table*}

\noindent\textbf{Ablation on Support Set.}~
To avoid overfitting and geometric ambiguity during single-view training of NeRF, we construct a support set for each view to provide multi-view constraints. Here, we further verify the effectiveness of the support set by removing this setting from our pipeline. As shown in Fig.~\ref{fig:abl_sprt}, the radiance field in experiment (a) is trained under the constraint of a single initial view, i.e., without support set constraints. Obviously, the NeRF is overfitting in the training view and cannot produce clear results in the inpainting view, which further leads to poor inpainted results. By contrast, the case with a support set achieves high-quality rendering results in the inpainting view. Accordingly, a clear and concordant inpainted result can be estimated by the pre-trained diffusion model. This is also reflected in the metrics in Tab.~\ref{tab:abla}. Additionally, we design a series of experiments to determine the hyper-parameters of the support set, including the number of warping views $\xi$ and shift distance $\zeta$. Specifically, we use different number of warping views and shift distance to conduct the support set and initialize the NeRF model. Then, we calculate the PSNR values within valid pixels between the rendered images $I_k^R$ and the DIBR-based warping results $I_k, M_k$ $\leftarrow$ $DIBR_{0\rightarrow i}$ to measure the quality of initialized NeRF: $psnr$ = $\frac{1}{N_t}\sum_{k=1}^{N_t} 10\log_{10}\left(\left\| \left(I_k^R - I_k \right) \odot M_k \right\|_2\right)$, where $N_t$ is the number of test poses. We generate 100 test poses using a generation method similar to the support set poses, i.e., randomly sample $\zeta$ in the range $[0.1, 0.4]$. As shown in Fig.~\ref{fig:curve}(a), as the number of warped views increases, the benefit brought by the support set tends to saturate. To this end, we choose $\xi=8$ warping views in the experiments to balance the computation cost and the training benefit of the support set. By changing the shift distance $\zeta$ of support sets, as shown in Fig.~\ref{fig:curve}(b), we find that $\zeta=0.2$ can make the support set achieve better performance than other parameters. Therefore, we set $\zeta=0.2$ in all of our experiments.

\noindent\textbf{Ablation on Depth Loss.}~
Furthermore, to validate the effect of our depth loss, we compare our $L_2$ depth loss with the case without depth constraint and other two regularized depth losses, a Gaussian negative log likelihood (GNLL) \cite{roessle2022dense} depth loss and a scale and shift invariant (SSI) \cite{sargent2023vq3d} depth loss. Without the guidance of depth information, as shown in Fig.~\ref{fig:abl_loss}(a), the radiance field fails to synthesize novel views with plausible geometry and tends to produce ambiguous depth values in the near and far areas. 
In contrast, GNLL and SSI losses have better constraining effect on near or far depth, as shown in Fig.~\ref{fig:abl_loss}(b)\&(c). Still, they fail to achieve satisfactory results because their constraints are weaker than our $L_2$ constraint (Fig.~\ref{fig:abl_loss}(d) and Tab.~\ref{tab:abla}).
In fact, the depth information after alignment is view-consistent with the whole generated 3D scene and can be directly seen as ground truth. In this case, a stricter objective function is more effective in constraining the generated scene than these flexible loss functions.

\noindent\textbf{Ablation on Depth Alignment.}~
Moreover, we conduct an ablation study on our two-stage depth alignment strategy. In Fig.~\ref{fig:abl_align}(a), we present an example of scene generation without depth alignment, revealing noticeable demarcation lines caused by depth dislocations across different views. To address this issue, we introduce a two-stage depth alignment strategy. In the global alignment stage, we mitigate scale and value differences between known and newly predicted depth maps by computing the average scaling score and depth offset. Fig.~\ref{fig:abl_align}(b) demonstrates the impact of global alignment, where the newly estimated depth values are pulled closer to the known depth map as a whole. However, due to the non-linear nature of depth estimation by a neural network, differences among pixels do not vary linearly. Consequently, demarcation lines persist even with global alignment. In contrast, the local depth alignment fine-tunes a pretrained neural network to reduce local differences among pixels.
Comparing Fig.~\ref{fig:abl_align}(a) and (c), we observe that local alignment partially brings unaligned depth maps closer in a non-linear manner. However, without global alignment, it is challenging to eliminate such differences entirely. Therefore, we employ a two-stage depth alignment strategy to achieve smoother and more harmonious transitions at the demarcation lines, as depicted in Fig.~\ref{fig:abl_align}(d) and Tab.~\ref{tab:abla}.

\section{Conclusion}
\label{sec:conclusion}
In this paper, we propose the Text2NeRF for generating a wide range of 3D scenes with complicated structures and high-fidelity textures purely from a text prompt. We first leverage a pre-trained text-image diffusion model to generate an initial scene content and adopt a pre-trained monocular depth estimation model to provide geometric prior. Then, we initialize the radiance field of the scene according to the above information and update the 3D scene based on the PIU strategy. To avoid overfitting and geometric ambiguity during view-by-view updating, we introduce support sets to provide multi-view constraints for single-view training in NeRF. Moreover, we adopt depth and transmittance losses along with the RGB loss to achieve depth-aware NeRF optimization and propose a two-stage depth alignment strategy to eliminate depth disparity estimated in different views. Thanks to all well-designed modules and objectives, our Text2NeRF achieves to generate photo-realistic diverse 3D scenes with complex geometric structures and fine-fidelity textures.

\begin{figure}[t]
  \centering
  \includegraphics[width=1\linewidth]{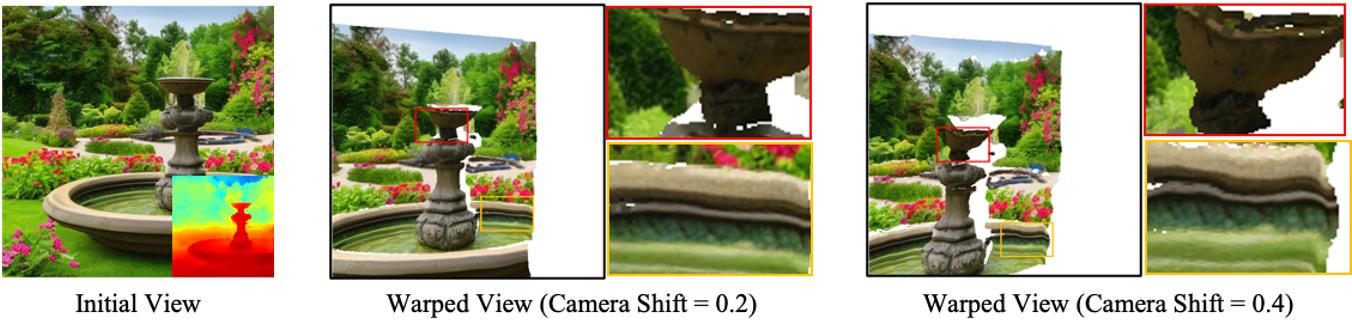}
  \caption{Geometric distortions caused by inaccurate depth estimation.}
  \label{fig:limi}
\end{figure}

\noindent\textbf{Limitation.} 
Although our scene generation experiments have yielded impressive results, it is important to acknowledge that our method struggles to generate scenes with very large occlusions due to the limited accuracy of the depth estimation. As illustrated in Fig.~\ref{fig:limi}, 
inaccurate depth estimation causes evident geometric distortion in the DIBR-based warped views, and this distortion becomes more pronounced with increased camera position offset. This results in noticeable artifacts and unrealism during the inpainting stage, which makes it difficult to generate reasonable results for our method. On the other hand, advancements in depth estimation techniques will effectively alleviate this limitation. 
Besides, as shown in Tab.~\ref{tab:exp_all}, compared to mesh or point cloud-based generation methods, our method, like the previous NeRF-based methods\cite{poole2022dreamfusion}, requires a longer optimization time (about 1.5 hours). Meanwhile, to generate 3D scenes in a large view range, we set the camera positions inside the radiance field and make the camera look outside. By this means, our method cannot generate an individual 3D object like other methods of setting the camera to look inside. To overcome this limitation, a flexible scene-adaptive camera setting strategy could be introduced in our framework in the future.

\bibliographystyle{IEEEtran}
\bibliography{IEEEabrv,ref}

\end{document}